\documentclass[sigconf]{acmart}
\usepackage{multirow}
\usepackage{booktabs} 
\usepackage{mathrsfs}
\usepackage{tabularx}
\DeclareMathOperator*{\argmax}{arg\,max}

\setcopyright{rightsretained}

\acmDOI{10.475/123_4}

\acmISBN{123-4567-24-567/08/06}

\acmConference[WWW 2018]{ACM Woodstock conference}{23 - 27 April 2018}{Lyon, France} 
\acmYear{1997}
\copyrightyear{2018}

\acmArticle{4}
\acmPrice{15.00}

\begin{document}
\title{Keyword-based Query Comprehending via Multiple Optimized-Demand Augmentation}

%

%
%
%
%

\author{Boyuan Pan$^*$, Hao Li$^*$, Zhou Zhao$^\dag$, Deng Cai$^*$, Xiaofei He$^*$}
\affiliation{
\institution{	$^*$State Key Lab of CAD$\&$CG, College of Computer Science, Zhejiang University, Hangzhou, China\\
	$^\dag$College of Computer Science, Zhejiang University, Hangzhou, China\\
	}
	$\{$panby, haolics, zhaozhou$\}$@zju.edu.cn, $\{$dengcai, xiaofeihe$\}$@gmail.com,  \\
}

\begin{abstract}
In this paper, we consider the problem of machine reading task when the questions are in the form of keywords, rather than natural language. In recent years, researchers have achieved significant success on machine reading comprehension tasks, such as SQuAD and TriviaQA. These datasets provide a natural language question sentence and a pre-selected passage, and the goal is to answer the question according to the passage. However, in the situation of interacting with machines by means of text, people are more likely to raise a query in form of several keywords rather than a complete sentence. The $keyword$-$based$ $query$ $comprehension$ is a new challenge, because small variations to a question may completely change its semantical information, thus yield different answers. In this paper, we propose a novel neural network system that consists a Demand Optimization Model based on a passage-attention neural machine translation and a Reader Model that can find the answer given the optimized question. The Demand Optimization Model optimizes the original query and output multiple reconstructed questions, then the Reader Model takes the new questions as input and locate the answers from the passage. To make predictions robust, an evaluation mechanism will score the reconstructed questions so the final answer strike a good balance between the quality of both the Demand Optimization Model and the Reader Model. Experimental results on several datasets show that our framework significantly improves multiple strong baselines on this challenging task.
\end{abstract}


\maketitle

\section{Introduction}
This paper proposes to tackle machine reading comprehension task when the query sentences are in the keyword form. When we communicate with machines by means of text, such as search engine or customer service chatbot, we usually don't type a complete natural sentence into the search box, because a whole query sentence may be too lengthy or sometimes hard to organize. For the sake of convenience, using a few keywords to represent the query becomes people's common choice. For instance, some users are more likely to type ``what obama did presidency'' when they are thinking about ``What did Barack Obama do during his presidency". In this case, people usually ignore the seemingly inconsequential words like  ``do" or ``during", but this may be notoriously ambiguous and can be interpreted multiple ways by machine for the task of reading comprehension\cite{bender2013linguistic} \cite{voorhees1999natural}. Moreover, there are  some cases where users find it hard to organize a short appropriate keyword-based query so they only select parts of the keywords to represent the demand, thus the resulting information incompletion is another tough challenge for machine comprehension.

\begin{figure}[t]
  \begin{center}
  \includegraphics[width=0.4 \textwidth]{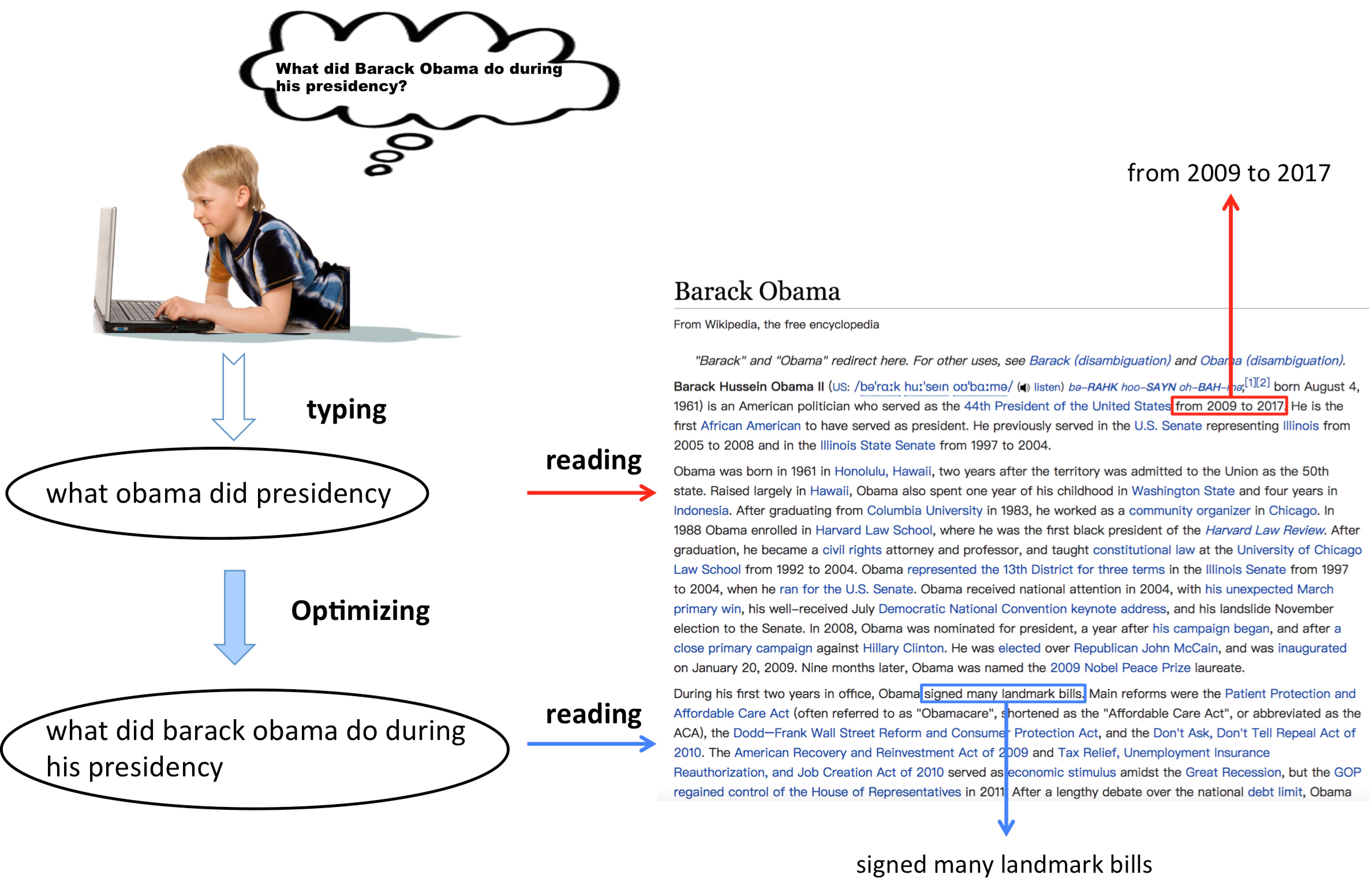}
  \caption{Keyword-based Query Comprehension.}
  \end{center}
\end{figure}

To overcome these troubles, many works have been done to provide natural language questions based on the keyword query to help disambiguate the demand\cite{Dror2013From}\cite{Zhao2011Automatically}\cite{37566}. However, most existing works are template-based methods, where the questions rules and patterns have to conform to existing keyword-question pairs. In this way, the generated questions are inherently limited to generalize to the wider structures that are not in previous queries and these methods barely have contribution on solving information incompletion mentioned above. Moreover, none of these works aim to the machine reading task, their evaluation metrics are those for text generation.

Machine reading comprehension is also a key problem in natural language processing. Recently, the performance on machine reading comprehension has been significantly boosted in the last few years with the introduction of large-scale machine reading datasets such as CNN/DailyMail \cite{hermann2015teaching}, SQuAD\cite{rajpurkar2016squad} and TriviaQA\cite{joshi2017triviaqa}. We only focus on the phrase-locating form task because it is relatively difficult and more practical in the real world. State-of-the-art systems\cite{seo2016bidirectional}\cite{pan2017memen}\cite{rnet} for these datasets usually encode the tokens in the document and question into word vectors from a lookup table and then employ a neural network sequence model that is combined with an attention mechanism. As is well-known that Long Short Term Memory(LSTM)\cite{hochreiter1997long} and Gated Recurrent Unit(GRU)\cite{cho2014learning} are skillful in handling time series and long text, but the keywords-based queries are short and may not have time continuity. Moreover, due to lack of syntactic structure, some augmentation for encoding layer in previous methods\cite{pan2017memen}\cite{liu2017structural} such as applying part-of-speech tags and name-of-entity tags can not be employed in this situation any more. To overcome the challenge of the $keyword$-$based$ $query$ $comprehension$, which is termed for this setting, we must make clear what the user's real demand is and provide a more readable question before locating the answer from the document.

In this paper, we present an end-to-end neural network system that combines three modules to comprehend and answer the queries. The first module called $Demand$ $Optimization$ $Model$, which takes the keyword-based query and the passage as input and then generates multiple optimized questions through a novel passage-attention neural machine translation. The second part of the system is the $Reader$ $Model$ that receives the new questions from the first module and predicts a probability distribution over the answers of each question. We also propose an Evaluation Mechanism along with the output part, which first evaluates the quality of the generated questions, then assigns the scores as weights to rank the answers for a best prediction. The contributions of this paper can be summarized as follows.

\begin{itemize}
\item First, we launch the $keyword$-$based$ $query$ $comprehension$ challenge and highlight its significance in the real-world applications with the limitations when employing existing machine reading comprehension methods on it. 

\item Second, we introduce a novel  neural network system to solve this task. Unlike previous machine reading methods, the query is first fed into a passage-attention neural machine translation and the final prediction is weighted by the quality of its corresponding new generated questions. 

\item Finally, we evaluate our model on two large machine reading datasets and achieve the state-of-the-art results on both of them compared to multiple competitive baselines.
\end{itemize}

We will discuss the Demand Optimization Model in detail in the section 2.1, the Reader Model in the section 2.2. The full question answering model and the Evaluation Mechanism is detailed in section 3.

\section{Framework}
Our model mainly consists of three parts. First, we input the keyword-based query and the passage into the Demand Optimization Model to generate the new candidate questions. Then, those questions and the passage are fed into a sequence model that is augmented with an attention mechanism. Finally, the Evaluation Mechanism scores each question as weights to their corresponding answers to obtain the prediction. In the following we will explain each module of the system.

\begin{figure*}[t]
  \begin{center}
  \includegraphics[height=0.27 \textheight]{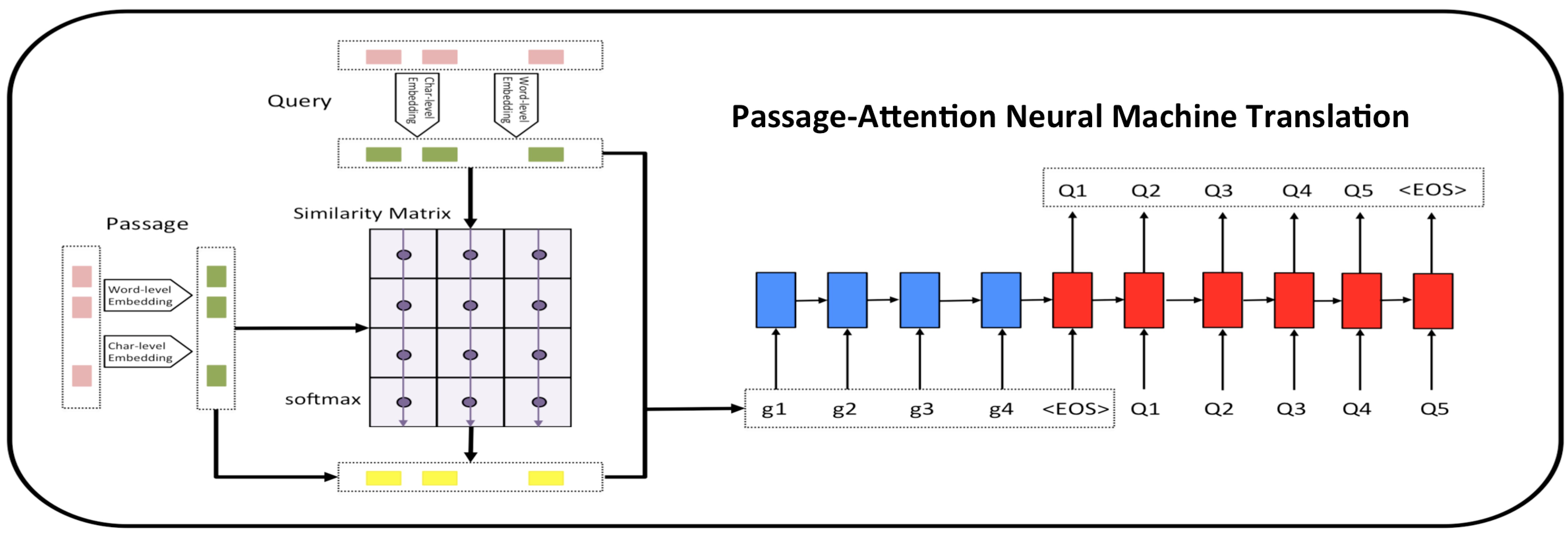}
  \caption{Overview of the Demand Optimization Model.}
  \end{center}
\end{figure*}

\subsection{Demand Optimization Model}
As figure 2 shows, we first introduce the Demand Optimization Model that based on the passage-attention neural machine translation. The model is inspired by the recent success of sequence-to-sequence learning in neural machine translation. Neural machine translation is trained end-to-end to maximize the conditional probability of a correct translation of a input sentence with a bilingual corpus. We can regard the keyword-based query sentence as source language and the ground truth question sentence as the other language. Moreover, we also input the passage to the process of the sequence to sequence as supplementary information to help optimize the query. 

The attributes of NMT offer advantages over traditional methods such as generating natural language without the need for specialized features, resources, or templates, and the NMT based approach has the potential of considering wider contextual information when generating text.

Firstly, we map each word to a vector space, in which word-level embeddings and character-level embeddings are applied. We use pre-trained word vectors \emph{GloVe}\citep{pennington2014glove} to obtain the fixed word embedding of each word. The character-level embeddings are generated by using Convolutional Neural Networks(CNN) which is applied to the characters of each word. This layer maps each token to a high dimensional vector space and is proved to be helpful in handling out-of-vocab(OOV) words\cite{Yang2016Words}.

We use a bi-directional long short term memory units(LSTM) to encode both the passage and keyword-based query embeddings and obtain their representations $\{ P_{j} \}^{n}_{j=1}$ and $\{ F_{i} \}^{m}_{i=1}$:

 $$F_{i} = {\rm BiLSTM}([w^{F}_{i}; c^{F}_{i}]), i \in [1,...,m]$$
 $$P_{j} = {\rm BiLSTM}([w^{P}_{j}; c^{P}_{j}]), j \in [1,...,n]$$
 
where $w$, $c$ represent word-level embedding and character-level embedding. $n$ and $m$ are the length of the passage and query respectively. The encoder recursively processes tokens one by one, and uses the encoded vectors to represent text.

Different from conventional sequence to sequence NMT, our passage-attention NMT contains the information of the passage, which is a significant basis for the optimizing process. We first obtain a similarity matrix by matching the passage embedding and the question embedding along with a softmax function:

$$s^j_i = \frac{{\rm exp}(match(F_i, P_j))}{\sum_{j^{\prime}}{\rm exp}(match(F_i, P_{j^{\prime}}))}$$
where 
$$match(x,y) = v_s^{\top} {\rm tanh}(W_s[x;y;x \circ y])$$
$v_s$ and $W_s$ are the weight parameters, $\circ$ is elementwise multiplication. Then a weighted passage based attention that represents which passage words are most relevant to each question word is computed:
$$a_i = \sum_j s^j_i P_j$$

After that, the question embeddings and the attention vectors are combined together to yield the input sequence:
$$g_i = W_g [F_i; F_i \circ a_i]$$
Then another bi-directional LSTM is employed by iterating the following equations:
$$h_i = {\rm BiLSTM}(g_i), i \in [1,...,m]$$

The decoder LSTM is subsequently applied to unfold the context vector $c_t$ into the target sequence, through the following dynamic process:
$$s_t = {\rm BiLSTM} (Q_t, c_t), t \in [1,...,T]$$
where T is the length of the target sentence, $Q_t$ is the output word, and:
$$\alpha_{ti} = \frac{{\rm exp}(match(s_{t-1}, h_i))}{\sum_{i^{\prime}}{\rm exp}(match(s_{t-1}, h_{i^{\prime}}))}$$
$$c_t = \sum_{i=1}^{m} \alpha_{ti} h_i$$

Since the optimized question is based on the input query, it would be better if the former can keep the important words from the latter. Inspired by Gu et al.\cite{Gu2016Incorporating}, we incorporate the copying mechanism into NMT, the probability of generating target word $y_t$ in the output sequence becomes:
$$P(Q_t|Q_{<t},F,P) = P_g(Q_t|Q_{<t},F,P) + P_c(Q_t|Q_{<t},F,P) $$
The first part $P_g(Q_t|Q_{<t},F,P) = \frac{1}{N}f'(s_t, Q_{t-1}, c_t)$, which is the probability of generating the term $Q_t$ from vocabulary. $f'$ denotes a softmax classifier and $N$ is the normalization term. The second part is the probability of copying it from the source sequence:
$$P_c(Q_t|Q_{<t},F,P) = \sum_{j:x_i = Q_t}{\rm exp}(\phi_c(x_i)), x_i \in \mathcal{X}$$
where $\mathcal{X}$ denotes all words in the source query sequence. We use beam search to decode tokens that maximizes the conditional probability. The results with the best decoding scores are considered candidate question sentences.

\begin{figure*}[t]
  \begin{center}
  \includegraphics[height=0.28 \textheight]{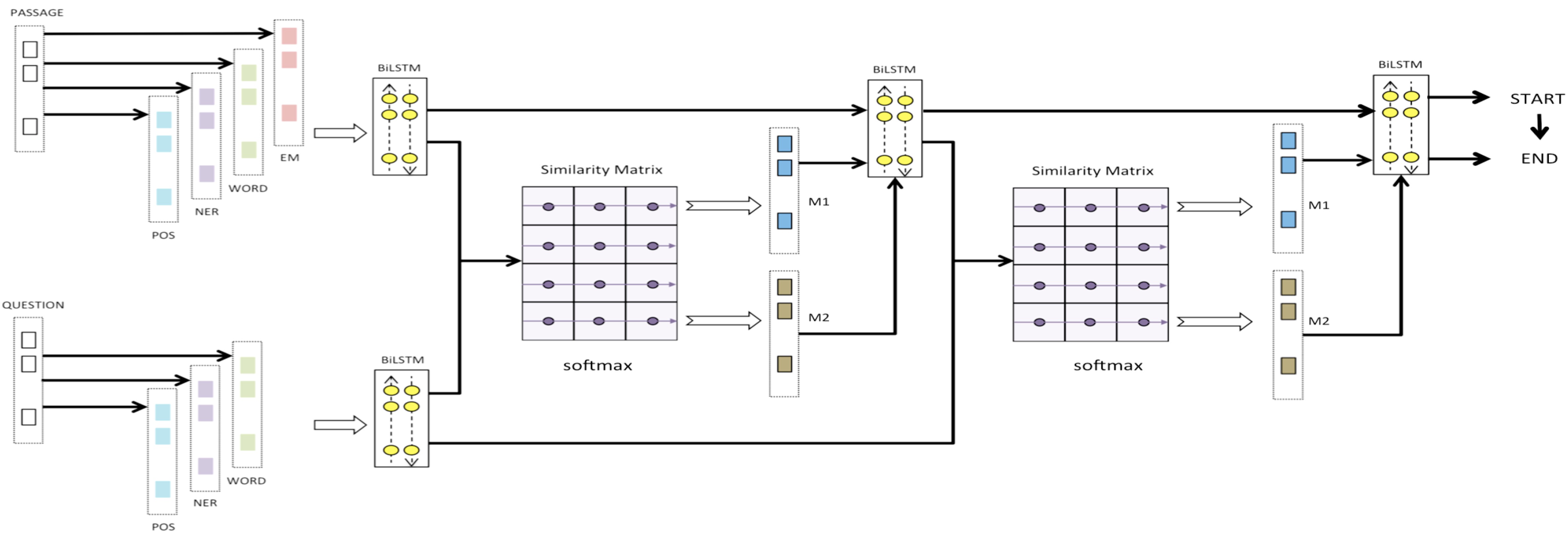}
  \caption{Overview of the memory network based Reader Model.}
  \end{center}
\end{figure*}

\subsection{Reader Model} 
In this section, we focus on the reading part where we assume that the question is a given natural language sentence. Recently, the neural network based models have achieved significant results on machine reading comprehension tasks. Our Reader Model is inspired by successful the models MEMEN\cite{pan2017memen} and Document Reader\cite{Chen2017Reading}.

\subsubsection{\textbf{Encoding Layer}}
As figure 3 shows, we first represent all tokens in the passage or the question as sequence of embeddings and pass them as the input to a recurrent neural network\cite{mikolov2010recurrent}. Similar to the encoding layer of Demand Optimization Model, we obtain word-level embeddings and character-level embeddings at beginning.

In addition, we improve the quality of embedding with implement of some features.
\begin{itemize}

\item To get syntactic information and semantic information of the words, we use the part-of-speech(POS) tags and named-entity recognition(NER) tags. We first create two new datasets by transforming all of the given training set into their POS tags and NER tags, in units of sentence. Then we employ the skip-gram model\cite{mikolov2013distributed} to the created datasets to obtain two new lookup tables, which represent POS tags embeddings and NER tags embeddings respectively. Finally, each word is able to get its own POS embedding and NER embedding by these lookup tables.

\item For every word in the passage, we want to know whether it is in the question sentence. We use three binary features to indicate whether the word can be exactly matched to any question word, which respectively means original form, lowercase and lemma form.

\end{itemize}

We use a bidirectional LSTM on top of the embeddings provided by the concatenation of features to model the temporal interactions between words and output the fixed-dimensional representation of passage and question, which are denoted by $\{ C_{i} \}^{n}_{i=1}$ and $\{ G_{j} \}^{l}_{j=1}$:

 $$C_{i} = {\rm BiLSTM}([w^{C}_{i}; c^{C}_{i}; s^{C}_{i}]), i \in [1,...,n]$$
 $$G_{j} = {\rm BiLSTM}([w^{G}_{j}; c^{G}_{j}; s^{G}_{j}]), j \in [1,...,l]$$
 
where $w$, $c$, $s$ represent word-level embedding, character-level embedding and the concatenation of feature vectors respectively.

\subsubsection{\textbf{Interaction Layer}}
We apply the structure of memory network and double orientations attention mechanism to enhance the effect, as figure 4 shows. We first obtain an alignment matrix $A \in R^{n \times l}$ between the query and context by $A_{ij} = v_{1}^{\top}[C_{i};G_{j};C_{i} \circ G_{j}]$, $v_1$ is the weight parameter. Now we use $A$ to obtain the attentions and the attended vectors in both orientations.

When we consider the relevance between context and query, the most representative word in the query sentence can be chosen by $e = {\rm max}_{row}(A) \in R^{n} $, where ${\rm max}_{row}(A)$ denotes the maximum member of all rows, and the attention is $d = {\rm softmax}(e)$. Then we obtain the first orientation attention vector:

$$m^{1} = \sum_{i} C_{i}  \cdot d_i$$

We tile $m^1$ for n times to get the attention matrix $M^1$ of the first orientation. For each passage word, there is an attention weight that represents the relevance degree to the query:

$$D = {\rm softmax}_{row} (A) \in R^{n \times l}$$
where ${\rm softmax}_{row}(A)$ means that the softmax function is performed across the row vector, and each attention vector is $M^{2}= D \cdot G $, which is based on the query embedding, hence the second orientation attention matrix is $M^2$. We use a simple linear function to concatenate the attention vectors and the original embedding of passage and then put them into a bi-directional LSTM. 

$$M = v^{\top}_{m}[M^1 \circ C; M^2 \circ C; M^2 - C; C]$$
where $v^{\top}_{m}$ is a training parameter. At last, we put $M$ into a bi-directional LSTM and get the output $\tilde{C}$ which captures the interaction among the passage words conditioned on the question. 

We regard $\tilde{C}$ as the input $\{ C_{i} \}^{n}_{i=1}$ of next layer in the memory network. In other words, the representation of passage and question in the next layer are denoted by $\{ \tilde{C}_{i} \}^{n}_{i=1}$ and $\{ G_{j} \}^{l}_{j=1}$, then repeat the process above and finally get the output $O$ after the bi-directional LSTM in the second layer of memory network.

\subsubsection{\textbf{Prediction Layer}}
In this layer, our model is required to predict the span of tokens that is most likely the correct answer. We take the $O$ and the passage embedding $C$ as input, and simply train two classifiers for predicting the start and end of the span:

$$P_{start} = {\rm softmax}(v^{\top}_{start}[C;O;C-O])$$
$$P_{end} = {\rm softmax}(v^{\top}_{end}[C;O';C-O'])$$
where $v^{\top}_{start}$ and $v^{\top}_{end}$ are trainable parameters, and $O'$ is the output of bidirectional LSTM whose input is $O$. For the loss function, we minimize the sum of the negative probabilities of the true start and end indices by the predicted distributions.

\section{Full Question Answering System}
\subsection{Pipeline}
In the full question answering system, we denote $F$ as the keyword-based query, $P$ as the passage and $a$ as its answer. The goal of our model is to compute the conditional probability $p(a|P,F)$ of the answer given the question and passage. We decompose it as following:

$$p(a|P,F) = \sum_{i}p(a|P,Q_i)p(Q_i | P, F)$$
where $p(a|P,Q_i)$ is obtain by the Reader Model, $p(Q_i | P, F)$ is obtained by the Demand Optimization Model and the following Evaluation Mechanism. $i$ is the number of questions the Demand Optimization Model generates. The details will be discussed in following subsections.

As is showed in Figure 4, we first employ the Demand Optimization Model to generate several candidate questions $Q_i$ from the given keyword-based query $K$ and passage $P$. Subsequently, the Reader Model combines the passage $P$ and each candidate question $Q_i$ to obtain the corresponding answer $a_i$. The Evaluation Mechanism will evaluate the quality of each optimized question $Q_i$ and assign them a score $s_i$ as the weight of the answer. In the end, the results of weight scores along with the distribution scores from the Reader Model are combined to predict the answer.

\begin{figure}[t]
  \begin{center}
  \includegraphics[width=0.45 \textwidth]{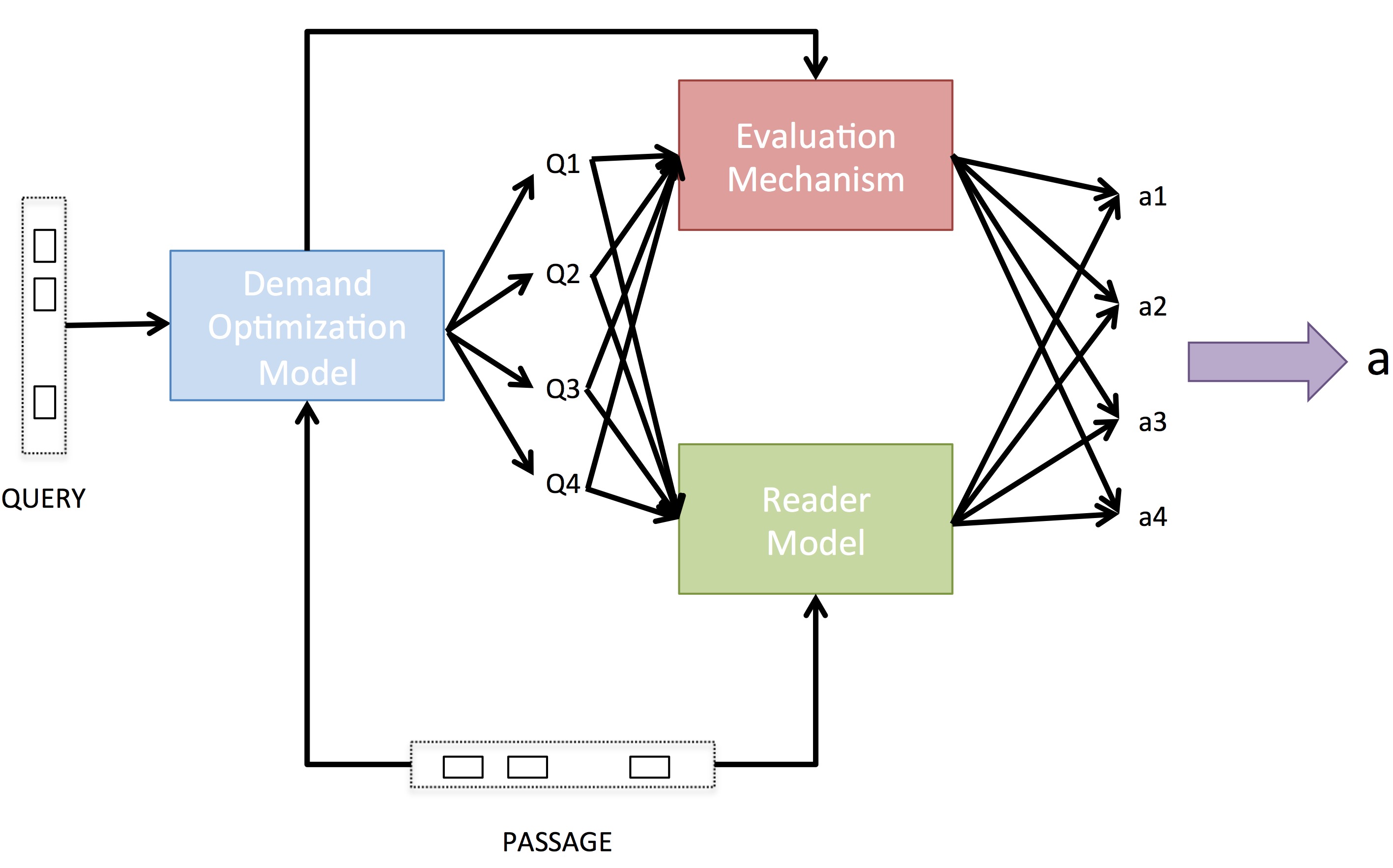}
  \caption{The pipeline of the full question answering system.}
  \end{center}
\end{figure}

\subsection{Evaluation Mechanism}
In order to estimate the generated question's quality $p(Q_i | P, F)$, our evaluation mechanism is based on neural networks, which in some perspective can also ensure the overlap between the keyword-based query and the generated question.

We first use $Glove$ to transform both the keyword-based query $F$ and the generated question $Q_i$ into vectors word by word and subsequently input them to a bi-directional LSTM. For each question, we represent it by the concatenation of the last hidden state in both direction of LSTM, denoted by $h_f$ and ${h_g}_i$ respectively. 

To compute the influence of $h_f$ to ${h_g}_i$ and obtain $p(Q_i | P, F)$, we use a linear function followed by a softmax over all the generated questions:

$$I_i = (v_h^{\top}[h_f; {h_g}_i; h_f \circ {h_g}_i; h_f - {h_g}_i])$$
$$p(Q_i | P, F) = \frac{{\rm exp}(I_i)}{\sum_{j \in \mathcal{Q}} {\rm exp}(I_j)}$$
where $v_h$ is a trainable parameter, $\mathcal{Q}$ is the set of generated questions.

\subsection{Training}
Before training the full question answering system, we pre-train the Demand Optimization Model using the ground-truth question as label. Our training object is formulated as follows:

$$J_{t} = -\sum_{(F,Q^g) \in \mathcal{D}}{\rm log}~p(Q^g|F,P)$$
where 
$${\rm log} ~p(Q^g|F,P) =  \sum_{t} {\rm log}~p(Q^g_t | Q^g_{<t},F,P)$$
where $\mathcal{D}$ is the set of all the <keyword-based, ground truth> question pairs in the dataset, $F$ is the keyword-based query, $Q^g$ is the ground truth question, and $P$ is the passage.

After the Demand Optimization Model is trained, we train the full question answer system end-to-end using the ground-truth answer as label. We employ a log-likelihood objective function for training which minimizes the following:

$$J = -\sum_{(P,F,a) \in \mathcal{E}}{\rm log} ~p(a|P,F)$$
where $\mathcal{E}$ is the set of all training data, $a$ is the answer and ${\rm log} ~p(a|P,F)$ is computed as showed before, which is $\sum_{i}p(a|P,Q_i)p(Q_i | P, F)$. For $p(a|P,Q_i)$, we compute the sum of the negative probabilities of the true start and end indices by the predicted distributions:

$$p(a|P,Q_i) = \frac{1}{N} \sum_{i}^{N}({\rm log}~P_{start_{i}} + {\rm log}~P_{end_{i}})$$
where $N$ is the size of the batch from the dataset $\mathcal{E}$, $P_{start_{i}}$ and $P_{end_{i}}$ are the true start and end position of the answer in the passage belong to the $i$-th example.

For the test time, we predict the answer by 
$$\tilde{a} =\argmax_{\hat{a} \in \mathcal{A}} p(\hat{a} | P,F) $$
where $\mathcal{A}$ is the set of candidate answers. 

\begin{table}
  \label{tab:commands}
  \begin{tabular}{ccl}
    \toprule
    \textbf{Dataset} &  \textbf{QA pairs} & \textbf{Paragraphs}\\
    \midrule
    SQuAD & 107785 & 23215 \\
    TriviaQA(Wikipedia) & 77582 & 138538\\ 
    TriviaQA(Web) & 95956 & 662659\\ 
    \bottomrule
  \end{tabular}
  \vspace{2mm}
  \caption{Datasets Statistics}       
\end{table}
 
\section{Data} 
Our work is based on two dataset, Stanford Question Answering Dataset (SQuAD) v1.1\cite{rajpurkar2016squad} and TriviaQA version 1.0\cite{joshi2017triviaqa}. Statistics of the datasets are given in Table 1. We make some modification to the questions in both datasets and the details are in the following.

\begin{table*}
    \label{tab:commands}
  \begin{tabular} {lp{8cm}p{7cm}p{7cm}}
    \toprule
    \textbf{Dataset} & \textbf{Original Question} & \textbf{Keyword-based Query}\\
    \midrule
    SQuAD & Who was this season's NFL MVP? & season NFL MVP \\
    &What kind of weapon did Tesla talk about? & kind weapon Tesla talk \\
    & To which technology type that Tesla worked on did the caption refer to? & technology type Tesla worked caption refer\\
    & The two AAA clubs divided the state into a northern and southern California as opposed to what point of view? & AAA divided state northern California opposed point view\\
     \midrule
   TriviaQA & Which element is mixed with gold to make red gold? &element mixed gold make red gold\\
   & What was the original use of the building which now houses the Tate Modern Art Gallery in London? & original use building houses Tate Modern Art London\\
    \bottomrule
  \end{tabular}
\vspace{2mm}
     \caption{Examples from the modified datasets. In each case we show a modified keyword-based query with its original question.}   
\end{table*}

\subsection{SQuAD}
SQuAD consists 100,000+ questions posed by crowdworkers on a set of Wikipedia articles, where the answer to each question is a segment of text from the corresponding passage, rather than a limited set of multiple choices or entities. Passages in the dataset are retrieved from English Wikipedia by means of Project Nayuki's Wikipedia's internal PageRanks. They sampled 536 articles uniformly at random with a wide range of topics, from musical celebrities to abstract concepts. The dataset is partitioned randomly into a training set(80\%), a development set(10\%), and a hidden test set(10\%). In our experiment, we use the development set as test set.

\subsection{TriviaQA}
TriviaQA is also a large and high quality dataset, it includes 95K question-answer pairs authored by trivia enthusiasts and independently gathered evidence documents, six per question on average, that provide high quality distant supervision for answering the questions. The crucial difference between TriviaQA and SQuAD is that TriviaQA questions have not been crowdsourced from pre-selected passages. Moreover, evidence set of TriviaQA consists of web documents, while SQuAD is limited to Wikipedia\cite{pan2017memen}. We transform the dataset to the format of SQuAD, where a question has only one corresponding passage.

\subsection{Modification}
We only modify the questions in the datasets, and keep the answers and passages as before. To simulate human's intuition, we randomly take out 95\% of the stopwords in the question sentence except some special words like ``when", ``where" that people won't ignore when they type their query.

Moreover, in consideration of the information loss in case of the fact that people hardly type a long sentence, we take out TF-IDF scores weighted words from small to large until the length of the sentence equal to 8 if the number of the words in stopword cleaned query is still larger than 8. In addition, we add some noise that 5\% of the pruning processes are random. Several examples are showed in Table 2.

\section{Experiment} 
\subsection{Implementation Settings}
The tokenizers we use in the step of preprocessing data are from Stanford CoreNLP \cite{manningstanford}. We also use part-of-speech tagger and named-entity recognition tagger in Stanford CoreNLP utilities to transform the passage and question. For the skip-gram model, our model refers to the \emph{word2vec} module in open source software library, \emph{Tensorflow}, the skip window is set as 2. The dataset we use to train the embedding of POS tags and NER tags are the training set given by SQuAD, in which all the sentences are tokenized and regrouped as a list. To improve the reliability and stability, we screen out the sentences whose length are shorter than 9. Generated questions are stemmed and lowercased, we put the top 6 candidate questions into the Reader Model. We augment questions with EOS symbols, which means end-of-sequence. Word vectors for these symbols are updated during training process. We use 100 one dimensional filters for CNN in the character level embedding, with width of 5 for each one. We set the hidden size as 100 for all the LSTM layers and apply dropout\cite{srivastava2014dropout} between layers with a dropout ratio as 0.2. We use the AdaDelta for optimization as described in \cite{zeiler2012adadelta}. For the memory networks, we set the number of hops as 2. 


 \begin{table}[t]
\centering
\label{my-label}
\begin{tabular}{lp{0.5cm}lp{1cm}lp{1cm}lp{1cm}lp{1cm}lp{1cm}lp{1cm}}
\hline
\multicolumn{1}{c}{\multirow{2}{*}{\textbf{Method}}} & \multicolumn{2}{c}{\textbf{SQuAD}} & \multicolumn{2}{l}{\textbf{TriviaQA(Wiki)}} & \multicolumn{2}{c}{\textbf{TriviaQA(Web)}} \\ \cline{2-7} 
\multicolumn{1}{c}{}                        & \textbf{EM}          & \textbf{F1}    & \textbf{EM}     & \textbf{F1}      & \textbf{EM}  & \textbf{F1}  \\ \hline
BiDAF                                       &  68.32 & 77.95       &         40.26           &   45.74                 & 41.08  & 47.40    \\
DrQA                                        & 69.44 & 78.26      &       42.23         &    46.32                & 42.55   &    47.98 \\
MEMEN                                       & \textbf{70.98} & 80.36       &           43.16         &  46.90                  &  \textbf{44.25}   &  48.34   \\
DORM                                        &  70.46 & \textbf{80.67}      &       \textbf{43.23}           &  \textbf{47.52}    &   44.14 &\textbf{49.02}   \\ 
\hline
\end{tabular}
  \vspace{2mm}
  \caption{Model performance on the SQuAD and TriviaQA. }       
\end{table}

\subsection{Evaluation Metrics}
Two metrics are utilized to evaluate model performance: Exact Match (EM) and F1 score. EM measures the percentage of the prediction that matches one of the ground truth answers exactly. F1 measures the overlap between the prediction and ground truth answers which takes the maximum F1 over all of the ground truth answers\cite{rnet}.

\subsection{Baselines}
We use three strong baselines of machine reading comprehension to prove the competitiveness of our model.\\
\textbf{BiDAF} : ~This model\cite{seo2016bidirectional} is to train a bi-directional attention flow to achieve a query-aware context representation for the task of machine comprehension.\\
\textbf{DrQA} : ~ This approach\cite{Chen2017Reading} uses a simple attention based model that incorporates many useful features in the encoding. There is a note that we only use the Document Reader part in DrQA because its retrieval part is for open domain QA task. \\
\textbf{MEMEN}: ~ This machine comprehension model\cite{pan2017memen}  is designed to employ a hierarchical orientations attention layer to help locating the answers.

\begin{table*}[t]
\centering
\label{my-label}
\begin{tabular}{lp{1cm}lp{1cm}lp{1cm}lp{1cm}lp{1cm}lp{1cm}lp{1cm}}
\hline
\multicolumn{1}{c}{\multirow{2}{*}{\textbf{Method}}} & \multicolumn{2}{c}{\textbf{SQuAD}} & \multicolumn{2}{l}{\textbf{TriviaQA(Wiki)}} & \multicolumn{2}{c}{\textbf{TriviaQA(Web)}} \\ \cline{2-7} 
\multicolumn{1}{c}{}                        & \textbf{EM}          & \textbf{F1}    & \textbf{EM}     & \textbf{F1}      & \textbf{EM}  & \textbf{F1}  \\ \hline
BiDAF                                       & 52.22       & 65.04       &         26.25           &   33.89                 & 27.54 &  35.26   \\
BiDAF(original)                             & 37.65       & 48.82       &         20.02       &  24.75               & 22.34    & 26.86    \\
DrQA                                        & 55.98       & 66.74       &          29.70           &   35.49                 & 31.53  &  36.99     \\
DrQA(original)                              & 39.47       & 50.34       &             21.78  &  26.98               & 22.76   & 28.73    \\
MEMEN                                       & 56.12       & 66.42       &      31.21              &  34.16                  & 31.94    &  35.32   \\
MEMEN(original)                             & 40.23       & 50.11       &      24.12              &    28.41                &       25.07       &    30.67 \\ \hline
DORM(without Demand Optimization Model)     & 56.53       & 67.15       &  30.70       &    35.76         &      33.09         &  37.25   \\
DORM(without Evaluation Mechanism)          & 58.76       & 69.01       &    33.02           &         37.64           & 35.01        & 38.96    \\
DORM(E2D)                                            & 59.23       & 69.89       &    35.14           &         39.55           & 38.11        & 41.80    \\
DORM                                        & \textbf{61.54}       & \textbf{72.62}  & \textbf{37.70}  & \textbf{41.54}   &  \textbf{39.14}  & \textbf{43.88 }   \\ \hline
\end{tabular}
  \vspace{2mm}
  \caption{Model performance on the modified SQuAD and TriviaQA. The word ``original" in the bracket means that the model is trained on the original dataset where the question is a natural language sentence.}       
\end{table*}

\begin{figure}[t]
  \begin{center}
  \includegraphics[width=0.5 \textwidth]{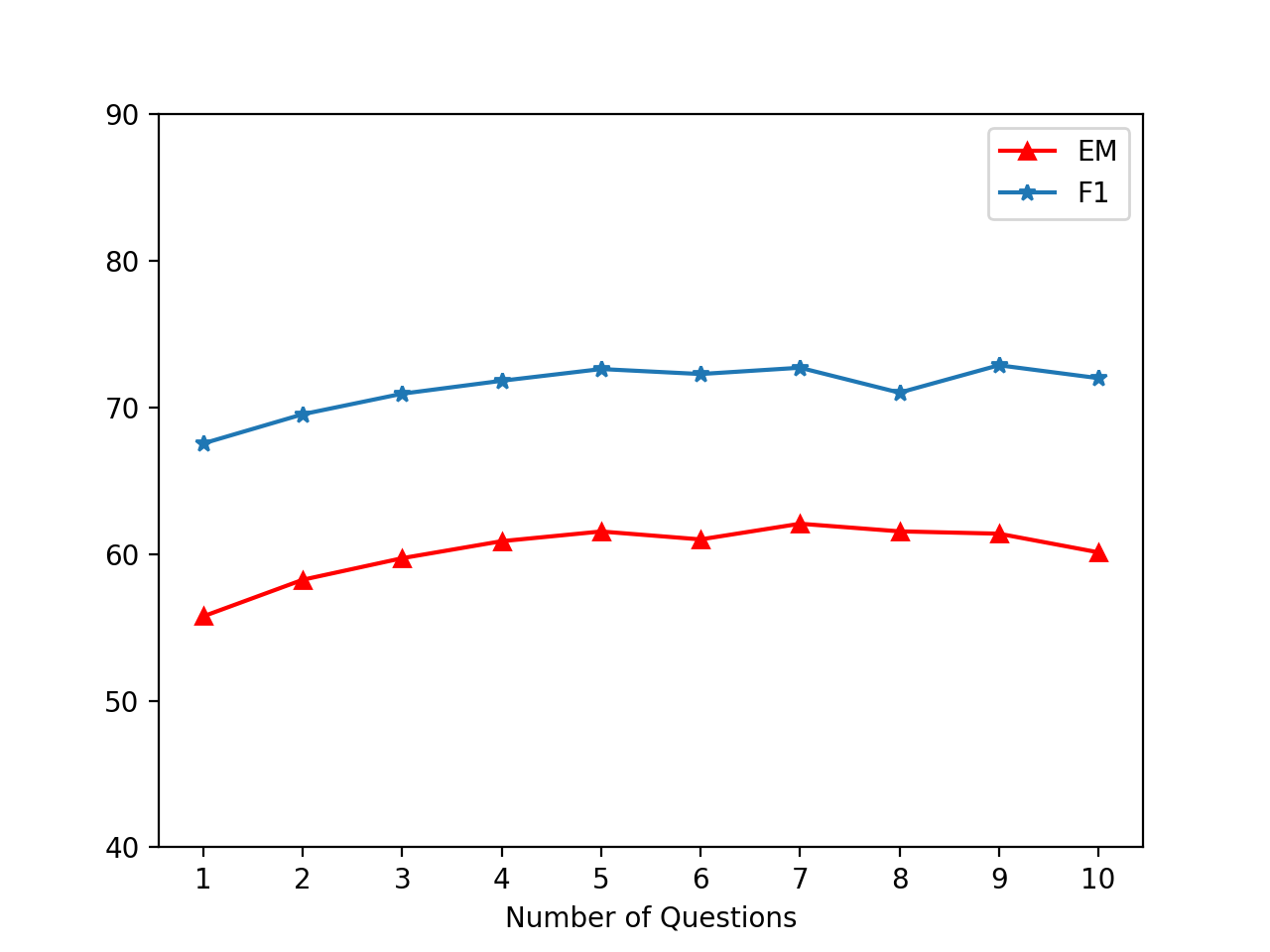}
  \caption{Performance on SQuAD with different number of generated questions.}
  \end{center}
\end{figure}

\subsection{Results}
To show the effectiveness of our method, we apply three competitive models BiDAF\cite{seo2016bidirectional}, DrQA\cite{Chen2017Reading}, MEMEN\cite{pan2017memen} as our baselines. There is a note that we only use the Document Reader part in DrQA because its retrieval part is for open domain QA task. 

We evaluate our model on the modified large scale dataset Stanford Question Answering Dataset (SQuAD) and TriviaQA, which consists two parts whose evidence documents are collected respectively from Wikipedia and the Web. As the Table 3 shows, we first test our Reader Model on the development set of the original datasets. The Reader Model in our model Demand-Optimization \& Reader Model(DORM) achieves the state-of-the-art result on the TriviaQA(Wikipedia), where the exact match is 43.23\%  and the F1 score is 47.52\%. Moreover, our model on other two datasets is also quite competitive among all the baselines. 

Table 4 presents our evaluation results on the modified SQuAD and TriviaQA. As we can see, our method clearly outperforms the all the baselines and achieves the state-of-the-art result. We observe that compared to their performance on the original datasets, the conventional machine reading comprehension models are barely satisfactory on the modified SQuAD and TriviaQA. Suffering from grammatical blunder and incomplete information, all the baselines decrease about 30\% in both of the EM and F1 scores on SQuAD and around 15\% on TriviaQA. However, when we trained these models on the modified training set, their performance improves dramatically, which is nearly 15\% higher than the models that trained on the original SQuAD training set and 8\% on TriviaQA training set. The reason might be that the neural network regards the keyword-based organization as a new kind of grammar.

Despite the huge progress from the change of the training data, the results on the new dataset still leave much to be desired. We conduct the ablations of our model to evaluate the individual contribution. We first run our model without the Demand Optimization Model, namely we only run the Reader Model on the test set, where the result is only a little better than the baselines above. Then we run all modules of the model but the Evaluation Mechanism, which means that the model generates questions but the final answer is only evaluated by the probability distribution in the Reader Model. We find that the exact match and F1 scores are around 2\% higher than the model without the Demand Optimization Model. This proves the utility of the Demand Optimization Model and the importance of natural language question reconstruction. However, the full question answering system is still much stronger. On the SQuAD, the complete DORM model achieve 61.54\% exact match and 72.62 F1 scores, which are nearly 3\% higher than the model without the Evaluation Mechanism. On the TriviaQA, the full question answering system is even 5 \% higher than the model without the Evaluation Mechanism. This result shows that the Evaluation Mechanism contribute towards the model performance, that is to say, the quality of the generated questions vary greatly and an evaluation system for it is necessary. We also use simple encoder-decoder(E2D) neural machine translation instead of Demand Optimization Model to generate questions, the results show that the information of passages which are incorporated our passage-attention NMT is very helpful.

\begin{figure*}[t]
  \begin{center}
  \includegraphics[height=0.4 \textheight]{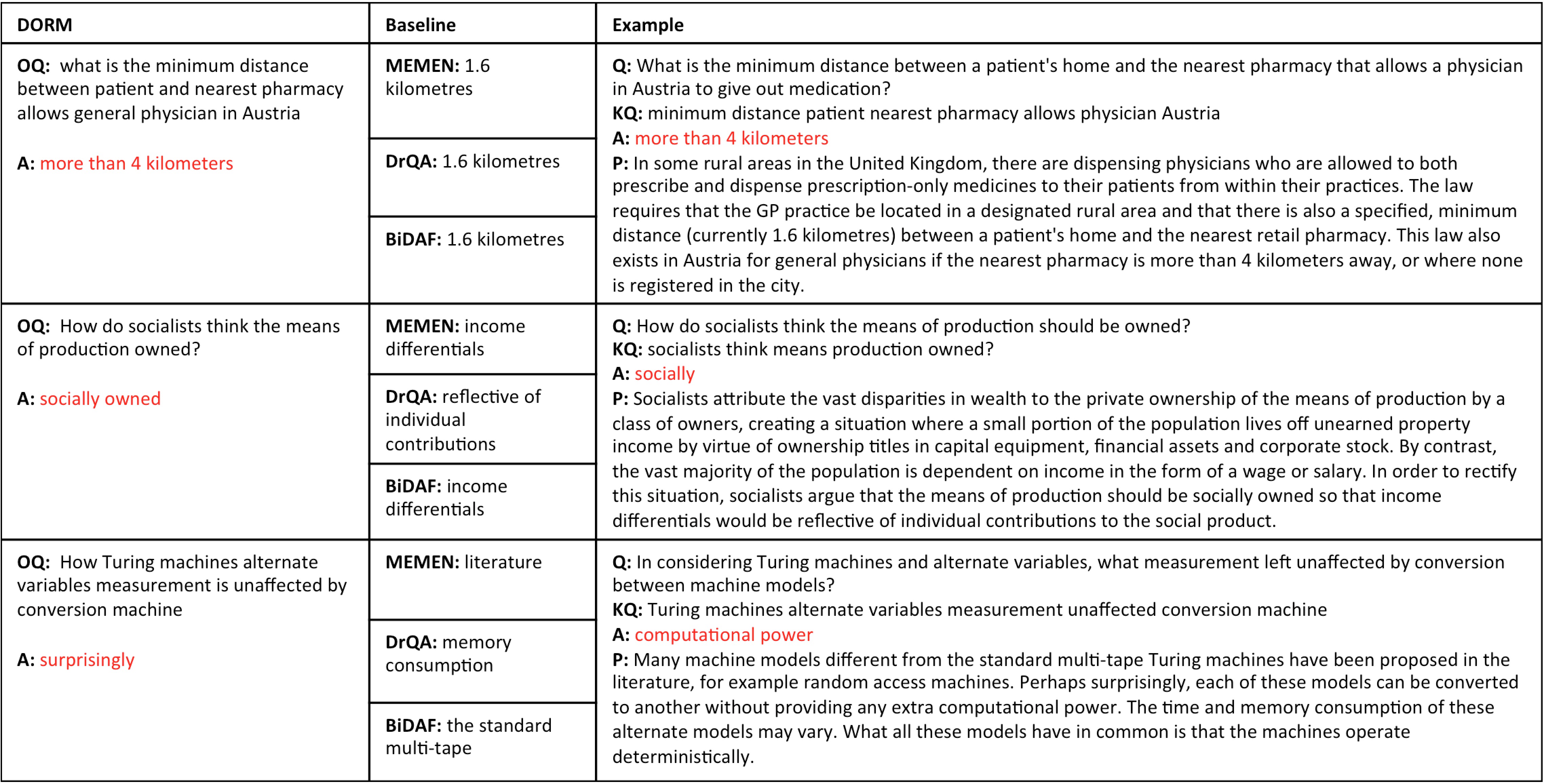}
  \caption{Examples of modified SQuAD with the optimized questions and the answer. The left column is the results of our model, OQ means optimized question. The medium column is the performance of the baselines. The right column is examples of the dataset, where KQ means keyword-based query.}
  \end{center}
\end{figure*}

\subsection{Answers and Questions Analysis}
We conduct some analysis of the results on dev set of SQuAD in order to better understand the behavior of our model.

Figure 5 shows the performance on SQuAD based on the change of the number of generated questions from the Demand Optimization Model. We find that the performance of the model tends to get better as the question number increases until it arrives 5. This reveals that various questions can efficiently help the model understand the demand. What is interesting is that when the number of questions is greater than 5, both of the exact match and F1 score begin to vibrate. We think this phenomenon is because that the real contributing generated questions are only the top 3 or top 5, too many options may be useless or even disturbing.

In the Figure 6, we show some examples of modified SQuAD with the optimized questions and the answer. We put our model's results in the left column, the baselines' results in the medium column and the examples in the right column. For each example, we denote the original question by Q, the keyword-based query by KQ, the answer by A, the passage by P, the optimized question by OQ. 

For the first instance, we can see that the question is in the situation of Austria, but the passage contains another possible answer which is in the situation of the United Kingdom. The keyword-based query contains the word ``Austria", but its chaotic structure might cause the reader model in the baselines pay more attention to understanding the front part. As a result, all of the baselines make clear most of the demand but ignore the important information ``Austria" thus give the wrong answer. In our model, we reconstruct the query and obtain a sentence that is more natural. The complementary word ``in" in front of ``Austria" seems to emphasize the last word and make the Reader Model realize the location information.

Different from the first case that both the original question and the keyword-based query are complex, the original question in the second instance is short and straightforward while the keyword-based query is a little confusing. This is the case where people may omit the stopword like ``how" when they type this kind of query. By means of keyword matching, all of the baselines find the right sentence, but they fail to understand the real demand. For our optimized question, we successfully predict the word ``how" so that the subsequent Reader Model is able to get the point. Although the final answer is not exactly right, the model generates a reliable question and provides a satisfactory answer.

In the last example, we present a typical fallible case from our model. As we can see, there is a comma symbol in the original question, which will usually not be typed into the command line by people and we took it out in the modified dataset. When people want to ask this kind of question, we often find it hard to organize a readable and short keyword-based query, and the KQ in the figure is the result sometimes people probably type. Moreover, the word ``what" is in the middle of the sentence, which is a difficult task for the Demand Optimization Model to recover. For our generated question, we predict the inquiry word ``how" at the beginning of the sentence, and totally construct a wrong structure compared to the original question. Since the neural network in the Reader Model is trained by a large scale of dataset, the answer to a question whose start word is ``how" has a large probability to be an adjective word or adverbial word. As a result, the predicted answer entirely makes no sense.

We also conduct a comparing experiment on the different type of questions. As we can see in Figure 7, our model performances best on the query whose original question contains ``when". This firstly because that the length of the answer of this kind of question is usually a short phrase or just a single word, moreover we didn't take the word ``when" out in the new dataset. On the other hand, questions contain ``why" tend to result in low accuracy, it is natural to realize that the explanatory answers are often hard to have perfect ground truths. In addition, since the limitation of training data, it is also a difficult task for machine to associate with the word ``why".

Besides the limitation of the model itself, the method to create the new dataset also has work to do to improve the utility. In many situations, we don't take out part of the phrase but using the abbreviation to replace it. The length of the query is also a point that is hard to handle, because our process will stop if the length of the sentence is less than 8, but in the reality people may continue to cut the words. Moreover, we keep the order of the original words in the question, but people sometimes would like to change it according to their own habits.

\begin{figure}[t]
  \begin{center}
  \includegraphics[width=0.5 \textwidth]{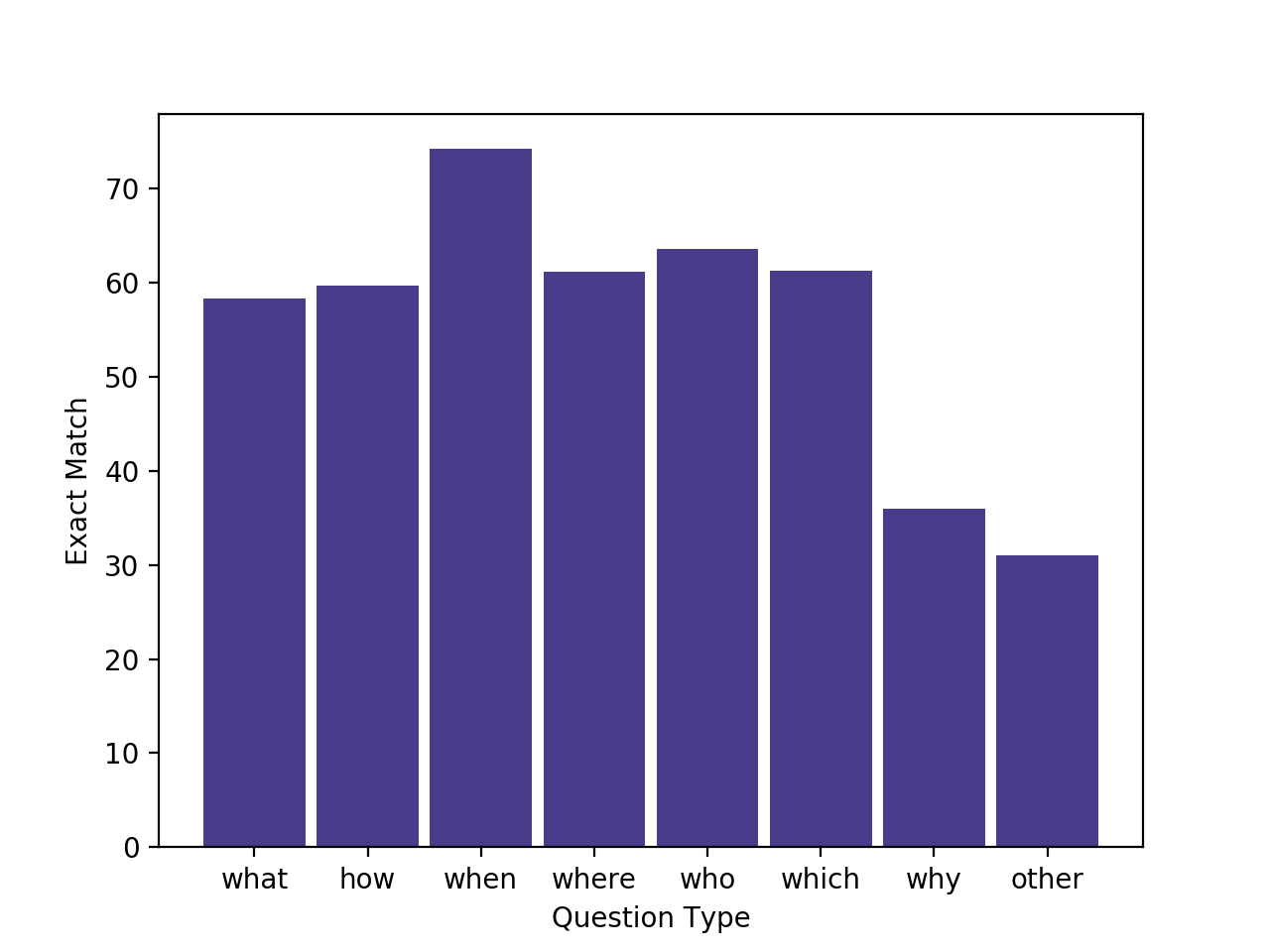}
  \caption{Exact match performance on SQuAD with different type of questions.}
  \end{center}
\end{figure}

\section{Related Work}
\subsection{Machine Reading Comprehension}
Several benchmark datasets play an important role in progress of machine comprehension research in recent years. The cloze-style datasets\citep{hill2015goldilocks}\cite{hermann2015teaching}\cite{onishi2016did}\cite{paperno2016lambada} are popular in last few years. However, these datasets are either not large enough to support deep neural network models or too easy to challenge natural language. Recently, Rajpurkar et al.\cite{rajpurkar2016squad} and Joshi et al.\cite{joshi2017triviaqa} released the Stanford Question Answering Dataset(SQuAD) and TriviaQA respectively. Different from cloze-style queries, the diversity of the answers and questions and the degree of syntactic divergence between the question and answer sentences make both of them high quality datasets. 

Many works are based on the task of machine reading comprehension, and neural attention models have been particularly 
successful\cite{xiong2016dynamic}\cite{cui2016attention}\cite{wang2016multi}\cite{seo2016bidirectional}\cite{Zhang2017Exploring}\cite{Gong2017Ruminating}. Xiong et al.\cite{xiong2016dynamic} present a coattention encoder and dynamic decoder to locate the answer. Cui et al.\cite{cui2016attention} propose a two side attention mechanism to compute the matching between the passage and query. Hu et al.\cite{hu2017mnemonic} propose self-aware representation and multi-hop query-sensitive pointer to predict the answer span. Shen et al.\cite{shen2016reasonet} propose iterarively inferring the answer with a dynamic number of steps trained with reinforcement learning. Wang et al.\cite{rnet} employ gated self-matching attention to obtain the relation between the question and passage.

In this paper, we propose a new challenge, the keyword-based query machine reading comprehension, which requires machine to firstly make clear what the real demand is and subsequently answer the question. Inspired by the above works, we present a novel framework that reconstructs the query and feeds multiple generated questions to a neural attention reader model, then the final answer combines both the quality of the answer and the quality of the question which is scored by an evaluation mechanism. For our reader model, we simplify the network structure and abandon the frequently-used self-matching in the attention mechanism while making a special effort to improve the encoding layer. 

\subsection{Question Generation}
Question generation\cite{Chali2015Towards}\cite{Labutov2015Deep}\cite{Yao2010Question}\cite{Heilman2011Automatic} draws a lot of attentions in recent years. QG is very necessary in real application as it is always time consuming to create large-scale QA datasets. Heilman et al.\cite{Heilman2011Automatic} proposed a overgenerate-and-rank framework consisting of three stages. They transform the statement to candidate questions by executing well-defined syntactic transformations after transforming a sentence into a simpler declarative statement . In the end, a ranker is used to select the high-quality questions. There are also some existing works that generates questions from knowledge base\cite{Song2016Domain}\cite{Serban2016Generating}. For instance, Serban et al.\cite{Serban2016Generating} proposed a neural network approach which takes a knowledge fact (including a subject, an object, and a predicate) as input while generating the question with a recurrent neural network.

Recent studies also investigate question generation for the reading comprehension and question answering task\cite{Zhou2017Neural}\cite{Du2017Learning}\cite{Tang2017Question}. Their approaches are typically based on the encoder-decoder framework, which could be conventionally learned in an end-to-end way. As the answer is a text span from the passage, it is helpful to incorporate the position of the answer span. However, those works generate questions only from the passage, and they mainly focus on the quality of the generated sentences. In this paper, we reconstruct the questions where the input are both the keyword-based query and the passage. Moreover, the only evaluation criteria is the accuracy of the question answering task, and whether the generated questions are exactly natural language sentences is less important.

\section{Conclusions}
In this paper, we introduce a new challenge, $keyword$-$based$ $query$ $comprehension$, and propose a novel framework to improve the performance on it. Unlike existing works that directly read the query with the passage, our Demand Optimization Model firstly generates multiple new questions to better represent the real demand. Our novel Evaluation Mechanism can also score the quality of each question thus the final answer combines both the quality of the answers and the quality of the constructed questions. In the future, we plan to improve the combination part of the generating model the quality of the keyword-based query dataset.

%
%

\bibliographystyle{ACM-Reference-Format}
\bibliography{keyword_reading} 

\end{document}